%% file: main.tex
\documentclass[11pt]{article}

\usepackage{acl}

\usepackage{mathptmx}
\usepackage{courier}
\usepackage{latexsym}
\usepackage[T1]{fontenc}
\usepackage[utf8]{inputenc}
\usepackage{microtype}
\usepackage{graphicx}
\usepackage{booktabs}
\usepackage{url}
\usepackage{hyperref}
\usepackage{tikz}
\usepackage{fontawesome5}
\usepackage{xcolor}       
\usetikzlibrary{shapes.geometric, arrows.meta, positioning}
\usepackage{pifont}

\hypersetup{
    colorlinks=true,
    urlcolor=blue,
    linkcolor=blue
}

\title{MudawanSn: A Gold-Standard Wolof--Arabic Parallel Corpus for Machine Translation}

\author{
  Mouhamed Mbaye \\
  NLP Engineer\\
  GalsenAI Lab \\
  \texttt{mouhamed.mbaye@galsen.ai}
  \And
  Thierno Diop \\
  Professeur\\
  Ministère de l'Éducation Nationale du Sénégal \\
  \texttt{thiernodiop0@gmail.com}
}

\date{}

\begin{document}
\maketitle

\vspace{0.1cm}
\begin{center}
    \footnotesize
    \raisebox{-0.25em}{\includegraphics[height=1.2em]{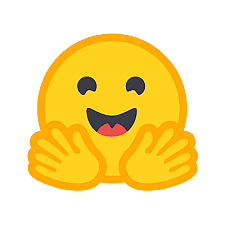}}~\textcolor{blue}{}
    \href{https://huggingface.co/datasets/mbaye930/wolof-arabic-parallel-corpus}{\nolinkurl{hf.co/datasets/mbaye930/wolof-arabic-parallel-corpus}} \\[3pt]
    \raisebox{-0.15em}{\faGithub}~\textcolor{blue}{}
    \href{https://github.com/M-mbaye30/MudawanSn}{\nolinkurl{github.com/M-mbaye30/MudawanSn}}
\end{center}
\vspace{0.2cm}

\begin{abstract}
We present MudawanSn, a gold-standard resource of 1,271 sentence-aligned pairs manually translated from Wolof into Modern Standard Arabic (MSA). The source texts are drawn from the MasakhaNER corpus \citep{adelani-etal-2022-masakhaner}, covering politics, society, religion, and sports in Senegalese news discourse. Although multilingual resources such as FLORES-200 \citep{nllbteam2024scaling} and NTREX \citep{federmann-etal-2022-ntrex} contain both Wolof and Arabic translations, to the best of our knowledge no publicly available parallel corpus has been specifically designed for the Wolof–Modern Standard Arabic language pair. We describe the corpus construction protocol, sentence alignment procedure, and quality-control workflow. We benchmark four machine translation systems spanning three architectural families: NLLB-200 (600M), mT5-base, and two AfriNLLB variants, showing that fine-tuning on MudawanSn yields substantial improvements in both translation directions. The best-performing model, AfriNLLB-12, achieves 7.76 BLEU and 30.72 chrF++ for Wolof-to-Arabic, and 8.75 BLEU and 33.08 chrF++ for Arabic-to-Wolof. The corpus is released under CC BY-NC and publicly available on Hugging Face and GitHub.
\end{abstract}

\footnotetext[1]{\url{https://github.com/masakhane-io/masakhane-ner/}} 
\footnotetext[2]{\url{https://github.com/AfriNLP/AfriNLLB}}

\input{sections/introduction}
\input{sections/related_work}

\input{sections/corpus_construction}

\input{sections/baseline}

\input{sections/conclusion}

\input{sections/limitation}
\bibliography{references}

\end{document}

%% file: sections/introduction.tex
\section{Introduction}
\label{sec:intro}

Wolof is the principal language of daily communication for approximately ten million speakers across Senegal, The Gambia, and Mauritania, where it coexists with the respective official languages in multilingual settings \citep{robert:hal-01513269}. Despite this large speaker population, Wolof remains severely under-represented in natural language processing research. The few existing parallel resources pair Wolof almost exclusively with French, reflecting Senegal's colonial and administrative context but leaving other culturally relevant language pairs entirely unaddressed.

The Wolof–Modern Standard Arabic (MSA) language pair is one such case. Senegal is a predominantly Muslim country with a long tradition of Arabic literacy through Qur'anic education, which led to the development of Wolofal, an Arabic-based Ajami writing system for Wolof \citep{Ajami_Scripts_Senegalese_2017}. Despite these historical and cultural ties, modern NLP resources for the Wolof–MSA language pair remain scarce. Although multilingual resources such as FLORES-200 \citep{nllbteam2024scaling}, NTREX \citep{federmann-etal-2022-ntrex}, and FineTranslations \citep{penedo2026finetranslations} include both Wolof and Arabic, they were not specifically designed as manually curated parallel corpora for this language pair. To the best of our knowledge, no publicly available manually translated, gold-standard parallel corpus has previously been developed for Wolof–MSA machine translation. Consequently, multilingual systems such as NLLB-200 \citep{nllbteam2024scaling} technically support both languages, yet our benchmark results indicate that their zero-shot translation quality on this language pair remains very limited.

To bridge this gap, we present MudawanSn, a publicly available Wolof–Arabic parallel corpus. Comprising 1,271 sentence-aligned pairs manually translated from Wolof into Modern Standard Arabic (MSA), our dataset is derived from Senegalese news articles in the MasakhaNER corpus ~\citep{adelani-etal-2022-masakhaner}. Our main contributions are as follows:

\begin{itemize}
  \item The release of MudawanSn, a publicly available gold-standard Wolof–MSA parallel corpus comprising 1,271 manually translated and sentence-aligned pairs, released together with comprehensive documentation and reproduction resources.

  \item A detailed and reproducible construction protocol for building a gold-standard parallel corpus for a previously under-resourced language pair. The protocol encompasses sentence extraction and translation, multilingual source alignment, cross-lingual semantic verification using LASER3 \citep{heffernan2022laser}, and rigorous quality-control procedures.

  \item A benchmark of four machine translation architectures, NLLB-200~\citep{nllbteam2024scaling}, mT5-base~\citep{xue-etal-2021-mt5}, and two AfriNLLB variants~\citep{moslem-etal-2026-afrinllb}, showing that fine-tuning substantially improves translation quality in both directions, with AfriNLLB-12 achieving the best overall performance (Section~\ref{sec:experiments}).
\end{itemize}

%% file: sections/related_work.tex
\section{Related Work}
\label{sec:related}

A data paper's contribution is best understood by situating it within the
existing landscape of language resources. We therefore review (i) parallel
corpora involving Wolof, (ii) multilingual resources that include both Wolof
and Arabic, and (iii) the specific gap addressed by MudawanSn.
Table~\ref{tab:related-resources} summarizes the principal resources discussed
in this section.

\begin{table*}[t]
\centering
{\footnotesize
\begin{tabular}{p{2.45cm}p{1.6cm}rccp{1.05cm}p{3.5cm}}
\toprule
\textbf{Dataset} &
\textbf{Pair(s)} &
\textbf{\# Pairs} &
\textbf{Pub.?} &
\textbf{Transl.} &
\textbf{Use} &
\textbf{Key characteristic} \\
\midrule

SENCORPUS \citep{nguer-etal-2020-sencorpus}
& Fr--Wo
& $\sim$70,000
& \ding{55}
& Human
& MT
& Closed; Fr--Wo only
\\

MAFAND-MT~\citep{adelani-etal-2022-thousand}
& Fr/En $\leftrightarrow$ Afr.\ (incl.\ Wo)
& 6,366
& \ding{51}
& Human
& MT
& No dedicated Wo--MSA pair
\\

FLORES-200~\citep{nllbteam2024scaling}
& Multiling.\ (incl.\ Wo, Ar)
& 2,024
& \ding{51}
& Human
& Eval
& Evaluation benchmark only
\\

NTREX-128~\citep{federmann-etal-2022-ntrex}
& Multiling.\ (incl.\ Wo, Ar)
& 1,997
& \ding{51}
& Human
& Eval
& Evaluation benchmark only
\\

SMOL~\citep{caswell-etal-2025-smol}
& Multiling.\ (incl.\ Wo)
& $\sim$97k
& \ding{51}
& Human
& Pretr.
& English-centric focus
\\

FineTranslations \citep{penedo2026finetranslations}
& Massively multiling.\ (Wo, Ar)
& 3.33B
& \ding{51}
& Auto.
& Pretr.
& Automatic; uncurated
\\

Mbaye et al.~\citep{Mbaye2023LowResourcedMT}
& Fr--Wo
& 123,000
& \ding{55}
& Human
& MT
& Closed; Fr--Wo only
\\

\midrule
\textbf{MudawanSn}
& \textbf{Wo--MSA}
& \textbf{1,271}
& \ding{51}
& \textbf{Human}
& \textbf{Res.+ bench.}
& \textbf{First public, human-curated, dedicated }
\\
\bottomrule
\end{tabular}
}
\caption{
Comparison of existing resources relevant to Wolof--Modern Standard Arabic
machine translation. ``Pub.?'' indicates unrestricted public availability;
``Transl.'' specifies whether translations are human- or machine-generated.
}
\label{tab:related-resources}
\end{table*}

\subsection{Parallel Corpora Involving Wolof}
\label{sec:wolof-resources}

Wolof has received increasing attention in NLP over the past few years,
leading to the release of several parallel and multilingual resources.
However, these resources predominantly pair Wolof with French or English,
or include it as one language among many in massively multilingual
collections. Table~\ref{tab:related-resources} summarizes the principal
resources discussed below.

\paragraph{French--Wolof.}
Early work on Wolof machine translation focused almost exclusively on
French. SENCORPUS~\citep{nguer-etal-2020-sencorpus} was among the first
French--Wolof parallel corpora developed for neural machine translation,
covering multiple domains including education, religion, legislation,
and society. MAFAND-MT~\citep{adelani-etal-2022-thousand} later introduced
a publicly available, manually translated Wolof subset of 6,366 sentence
pairs collected from four Senegalese news outlets. More recently,
Mbaye et al.~\citep{Mbaye2023LowResourcedMT} presented a substantially larger
French--Wolof corpus of 123,000 sentence pairs for neural machine
translation experiments. However, unlike MAFAND-MT, both SENCORPUS and the corpus of Mbaye et al. remain unavailable to the research community,
limiting reproducibility and downstream reuse.

\paragraph{English--Wolof.}
Wolof is also represented in multilingual evaluation benchmarks.
FLORES-200~\citep{nllbteam2024scaling} provides professionally translated
evaluation sets covering 200 languages, including both Wolof and Modern
Standard Arabic, while NTREX-128~\citep{federmann-etal-2022-ntrex}
contains professionally translated news test sets for 128 languages.
These benchmarks are intended for evaluation rather than model training
and therefore do not provide large-scale parallel corpora for developing
machine translation systems.

\paragraph{Massively multilingual resources.}
Recent multilingual initiatives have further expanded Wolof coverage.
NLLB-200~\citep{nllbteam2024scaling},
MADLAD-400~\citep{Kudugunta2023MADLAD400AM},
SMOL~\citep{caswell-etal-2025-smol},
and FineTranslations~\citep{penedo2026finetranslations}
all include Wolof among hundreds of languages.
These resources primarily target broad multilingual coverage and
large-scale model pretraining rather than the development of
high-quality parallel corpora for specific language pairs. In
particular, FineTranslations relies largely on automatically generated
translations rather than manually curated bilingual data.

\paragraph{What is missing.}
Despite these important advances, publicly available Wolof parallel
resources remain overwhelmingly centered on French or English, either
directly or through multilingual collections. To the best of our
knowledge, no publicly available, manually translated, gold-standard
parallel corpus has previously been developed specifically for the
Wolof--Modern Standard Arabic language pair. MudawanSn addresses this
gap by providing such a resource together with a reproducible
construction protocol and machine translation baselines.

\subsection{Parallel Corpora Involving Arabic and African Languages}

\label{sec:arabic-african}

Although Arabic is widely represented in multilingual machine translation benchmarks, dedicated parallel resources pairing Arabic with sub-Saharan African languages remain relatively scarce compared with English- or French-centered resources \citep{adelani-etal-2022-thousand,nllbteam2024scaling}.
The largest multilingual collections, including OPUS~\citep{tiedemann-2012-parallel}, NLLB-200~\citep{nllbteam2024scaling}, and MADLAD-400~\citep{Kudugunta2023MADLAD400AM}, include Arabic alongside numerous African languages.
However, many African--Arabic language pairs are supported through multilingual transfer or pivoting rather than dedicated bilingual parallel corpora \citep{nllbteam2024scaling,Kudugunta2023MADLAD400AM}.

Second, recent African-focused MT projects targeting Arabic, including the
ongoing AfriNLLB initiative~\citep{moslem-etal-2026-afrinllb}, address the resource scarcity problem at scale but do not yet provide curated, sentence-aligned, gold-standard corpora for specific African--Arabic
language pairs. This remains an important limitation because multilingual
transfer and zero-shot translation between low-resource language pairs
generally underperform systems fine-tuned on directly parallel data
\citep{JMLR:v22:20-1307, adelani-etal-2022-thousand}. This limitation is also
reflected in our experiments (Section~\ref{sec:experiments}), where
multilingual models exhibit very weak zero-shot performance before
fine-tuning on MudawanSn. The recent survey of Senegalese
NLP~\citep{sen-nlp-survey} likewise identifies the lack of publicly available
Wolof--Arabic resources as an important gap for future research.

\subsection{Positioning of Our Contribution}
\label{sec:positioning}

Our corpus addresses the intersection of the two gaps identified above.
To the best of our knowledge, it is the first publicly available,
human-curated parallel resource pairing Wolof directly with Modern
Standard Arabic at the sentence level for the news domain.

MudawanSn differs from previous resources in three respects. First, it is
\emph{gold-standard}: every sentence pair has been manually translated and manually verified, in contrast to the web-mined or pivoted Arabic--African data found in massively multilingual collections. Second, it is \emph{news-domain and culturally situated}: the source texts come from four Senegalese outlets covered by MAFAND-MT, anchoring the corpus in a register and a content universe that is directly relevant to the Wolof-speaking community. Third, it is \emph{released with full provenance}: each pair is traceable to its upstream source through the matching procedure described in Section~\ref{sec:construction}, supporting downstream studies that wish to control for source register or domain.

%% file: sections/corpus_construction.tex
\section{Corpus Construction and Quality Control}
\label{sec:construction}

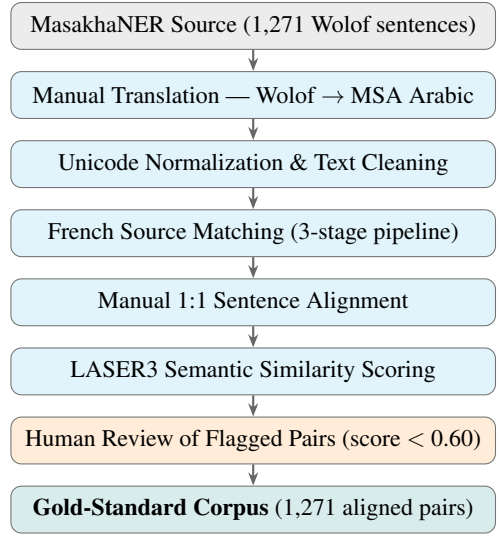
\begin{figure}[t]
\centering
\begin{tikzpicture}[
  node distance   = 0.28cm,
  box/.style      = {rectangle, rounded corners=4pt,
                     minimum width=6.4cm, minimum height=0.62cm,
                     text centered, draw=black!55, font=\small},
  srcbox/.style   = {box, fill=gray!15},
  procbox/.style  = {box, fill=cyan!10},
  qabox/.style    = {box, fill=orange!15},
  outbox/.style   = {box, fill=teal!15},
  arr/.style      = {-{Stealth[length=5pt]}, thick, draw=black!60}
]
\node[srcbox]  (src)    {MasakhaNER Source (1{,}271 Wolof sentences)};
\node[procbox, below=of src]    (trans)  {Manual Translation --- Wolof $\to$ MSA Arabic};
\node[procbox, below=of trans]  (clean)  {Unicode Normalization \& Text Cleaning};
\node[procbox, below=of clean]  (match)  {French Source Matching (3-stage pipeline)};
\node[procbox, below=of match]  (align)  {Manual 1:1 Sentence Alignment};
\node[procbox, below=of align]  (laser)  {LASER3 Semantic Similarity Scoring};
\node[qabox,   below=of laser]  (review) {Human Review of Flagged Pairs (score $<$ 0.60)};
\node[outbox,  below=of review] (gold)   {\textbf{Gold-Standard Corpus} (1{,}271 aligned pairs)};
\foreach \from/\to in {src/trans,trans/clean,clean/match,match/align,align/laser,laser/review,review/gold}
  \draw[arr] (\from) -- (\to);
\end{tikzpicture}
\caption{End-to-end pipeline from raw MasakhaNER source texts to the verified gold-standard Wolof--Arabic parallel corpus.}
\label{fig:pipeline}
\end{figure}

Building a high-quality parallel corpus for a language pair with no prior bilingual resources requires a multi-stage pipeline ensuring translation fidelity and alignment correctness. We therefore designed and executed an end-to-end process to clean, enrich, align, and semantically verify the manually translated sentence pairs, resulting in a verified gold-standard resource. Figure~\ref{fig:pipeline} provides an overview of the full pipeline.

\subsection{Source Data and Provenance}
\label{sec:sources}

The Wolof source sentences were selected from the MasakhaNER corpus \citep{adelani-etal-2022-masakhaner}, which aggregates news articles from four major Senegalese online outlets: \textit{Seneweb}, \textit{Jotna News}, \textit{YerimPost}, and \textit{SocialNetLink}. These texts represent contemporary written Wolof and cover diverse thematic areas such as politics, society, religion, and sports in Senegalese public discourse.

We selected MasakhaNER as our source for three scientific reasons. First, the texts were authored by native Senegalese speakers, ensuring authentic and natural Wolof news registers. Second, the corpus is released under a CC BY-NC 4.0 license, allowing the redistribution of derivative translated resources under compatible terms. Third, the sentences had already undergone high-quality segmentation and basic preprocessing, providing a clean foundation for cross-lingual alignment. From these source texts, a subset of 1{,}271 sentences was selected for the translation pipeline.

\subsection{Manual Translation}
\label{sec:translation}

The Arabic portion of the corpus was translated directly from the Wolof source by a bilingual native speaker of Wolof with advanced competence in Modern Standard Arabic (MSA). The translation followed a hybrid literal-semantic strategy: literal translation was preferred whenever it yielded a grammatical and idiomatic Arabic sentence, while semantic (meaning-based) translation was applied when a literal rendering would have produced unidiomatic or ambiguous Arabic. 

To ensure maximum accuracy, the translator relied on the monolingual Arabic dictionary Almaany\footnote{\url{https://www.almaany.com}} to verify word senses and registers, and consulted with native Wolof and Arabophone speakers when translating complex cultural or idiomatic concepts. No commercial machine translation systems were used in the translation loop, avoiding any circularity bias for downstream MT evaluation.

\subsection{Cleaning and Normalization}
\label{sec:cleaning}

We designed and executed a uniform preprocessing pipeline to clean the raw bilingual texts. We applied {Unicode Normalization Form C (NFC)} to both the Wolof and Arabic texts to resolve decomposed diacritics and composite characters. We programmatically stripped non-standard whitespaces (e.g., non-breaking spaces) and collapsed multiple whitespace sequences into a single standard space. Punctuation marks were standardized, and any residual formatting artifacts were removed to ensure a clean, unified textual representation.

\subsection{Multilingual Source Matching and Enrichment}
\label{sec:enrichment}

To enrich our corpus and enable future multi-way translation studies, we designed a custom three-stage matching pipeline to pair our Wolof sentences against the larger \texttt{galsenai/french-wolof-translation} \footnote{\url{https://huggingface.co/datasets/galsenai/french-wolof-translation}} dataset, thereby recovering the original French source sentences and metadata:
\begin{enumerate}
    \item \textbf{Exact Normalized Match:} We lowercased both strings, removed punctuation, collapsed spaces, and compared them. This yielded an exact match rate of 53.89\% (685 sentences).
    \item \textbf{Substring Match:} For source sentences that had been segmented into multiple shorter sentence fragments, we checked if the normalized Wolof sentence was a substring of any reference sentence, recovering an additional 34.30\% (436 sentences).
    \item \textbf{Fuzzy Token-Overlap Match:} For sentences with minor spelling variations, we calculated the token-level intersection. A match was established if at least 85\% of the words overlapped, recovering 9.99\% (127 sentences).
\end{enumerate}
Through this strategy, we successfully matched and enriched 98.19\% (1,248 out of 1,271) of our parallel pairs with their corresponding original French source sentences (\texttt{french\_source}) and source metadata (\texttt{source\_dataset} like MAFAND). The remaining 1.81\% are retained with MasakhaNER fallback records.

\subsection{Sentence Alignment and Quality Control}
\label{sec:alignment}

Each of the 1{,}271 sentence pairs was manually inspected to ensure strict one-to-one semantic correspondence, with no omissions or translation drift.

To quantitatively validate the parallel alignment and verify translation quality, we computed cross-lingual semantic similarity scores using the LASER3 model \citep{heffernan2022laser} as our primary quantitative check of alignment quality.

The similarity scores across the 1{,}271 parallel sentence pairs show high parallel fidelity, yielding a mean score of 0.7508 and a median score of 0.7649. Table~\ref{tab:similarity-dist} reports the detailed distribution.

\begin{table}[t]
\centering
{\small
\begin{tabular}{lr}
\toprule
\textbf{LASER3 cosine interval} & \textbf{Sentence pairs (\%)} \\
\midrule
$[0.0,\,0.4)$ & 0.31 \\
$[0.4,\,0.6)$ & 3.62 \\
$[0.6,\,0.8)$ & 69.47 \\
$\geq 0.8$    & 26.59 \\
\bottomrule
\end{tabular}
}
\caption{Distribution of LASER3 cross-lingual similarity scores over the 1,271 sentence pairs.}
\label{tab:similarity-dist}
\end{table}

This semantic evaluation supports the robust alignment and high quality of our dataset, with 96.06\% of all parallel sentence pairs scoring above 0.60 (and 26.59\% scoring $\ge 0.80$). All sentence pairs scoring below 0.60 were flagged and subjected to a rigorous manual review, correcting any alignment shifts, translation omissions, or spelling variations. The final corpus is released in a standard parallel format 
{\footnotesize\texttt{\href{https://huggingface.co/datasets/mbaye930/wolof-arabic-parallel-corpus}{tsv.}}}

\paragraph{Corpus Statistics.} To prevent the over-fragmentation bias typical of multilingual subword tokenizers (e.g., the SentencePiece tokenizers in NLLB or mT5, which frequently over-segment low-resource languages like Wolof), all corpus statistics are reported using simple whitespace tokenization (space-delimited words). Under this space-separated count, the 1{,}271 sentence pairs contain a total of 25{,}543 Wolof words and 17{,}972 Arabic words, with mean sentence lengths of 20.1 words (Wolof) and 14.1 words (Arabic). The vocabulary comprises 5{,}321 unique Wolof word types and 8{,}386 Arabic word types; the larger Arabic vocabulary size reflects the morphological richness of MSA. Source texts are predominantly Senegalese news from the MAFAND collection (98.2\%), with a small MasakhaNER fallback component (1.8\%).

%% file: sections/baseline.tex

\section{Baseline Experiments}
\label{sec:experiments}

\begin{table*}[t]
\centering
{\small
\begin{tabular}{llrrr}
\toprule
\textbf{Direction} & \textbf{Model} & \textbf{BLEU} & \textbf{chrF++} & \textbf{AfriCOMET\textsubscript{scaled}} \\
\midrule
wo $\rightarrow$ ar
 & NLLB-200 (600M) zero-shot       &  1.12 & 15.42 & $-12.22$ \\
 & NLLB-200 (600M) fine-tuned      &  6.47 & 29.31 & $\phantom{-}\mathbf{0.19}$ \\
\cmidrule{2-5}
 & mT5-base zero-shot              &  0.04 &  0.17 & $-40.43$ \\
 & mT5-base fine-tuned             &  0.43 & 11.10 & $-31.98$ \\
\cmidrule{2-5}
 & AfriNLLB-12enc-12dec zero-shot  &  3.97 & 25.39 & $-28.38$ \\
 & AfriNLLB-12enc-12dec fine-tuned & \textbf{7.76} & \textbf{30.72} & $-24.47$ \\
\cmidrule{2-5}
 & AfriNLLB-8enc-8dec zero-shot    &  4.53 & 25.53 & $-29.86$ \\
 & AfriNLLB-8enc-8dec fine-tuned   &  7.05 & 28.40 & $-28.66$ \\
\midrule
ar $\rightarrow$ wo
 & NLLB-200 (600M) zero-shot       &  3.14 & 18.24 & $\phantom{-}36.49$ \\
 & NLLB-200 (600M) fine-tuned      &  7.28 & 30.87 & $\phantom{-}\mathbf{41.93}$ \\
\cmidrule{2-5}
 & mT5-base zero-shot              &  0.02 &  2.57 & $-41.11$ \\
 & mT5-base fine-tuned             &  0.09 &  2.50 & $-50.83$ \\
\cmidrule{2-5}
 & AfriNLLB-12enc-12dec zero-shot  &  3.08 & 21.09 & $-14.71$ \\
 & AfriNLLB-12enc-12dec fine-tuned & \textbf{8.75} & \textbf{33.08} & $-3.18$ \\
\cmidrule{2-5}
 & AfriNLLB-8enc-8dec zero-shot    &  5.66 & 26.31 & $-7.49$ \\
 & AfriNLLB-8enc-8dec fine-tuned   &  8.18 & 31.95 & $-1.53$ \\
\bottomrule
\end{tabular}
}
\caption{Baseline machine translation results on the test split. Best score per direction and metric in bold.}
\label{tab:main-results}
\end{table*}

We assess the value of the proposed corpus along two axes: the magnitude of
fine-tuning gains it enables over zero-shot baselines, and the relative
behaviour of model families with different degrees of African-language
coverage. To this end, we evaluate four systems representing three
architectural families.

\subsection{Experimental Setup}

We split the $1{,}271$ sentence pairs into training ($85\%$, $n = 1{,}079$),
development ($7.5\%$, $n = 96$), and test ($7.5\%$, $n = 96$) sets. Splits
are released alongside the corpus to ensure reproducible evaluation.

\paragraph{Models.}
NLLB-200 distilled 600M~\citep{nllbteam2024scaling} is a massively
multilingual encoder--decoder MT model with explicit Wolof
(\texttt{wol\_Latn}) and Arabic (\texttt{arb\_Arab}) language codes.
mT5-base~\citep{xue-etal-2021-mt5} is a generalist text-to-text model
pre-trained on the multilingual mC4 corpus; we include it as an
out-of-distribution lower-bound reference, since mC4 contains substantial
Arabic but only marginal Wolof coverage.
AfriNLLB~\citep{moslem-etal-2026-afrinllb} is an
African-language-specialized variant of NLLB-200. We evaluate two
configurations: the full 12-encoder/12-decoder model and a pruned
8-encoder/8-decoder variant.

\paragraph{Training regimes.} Each model is evaluated in two settings.
Zero-shot: the off-the-shelf model is used directly on the test set
in both directions. Fine-tuned: we perform full-parameter
bidirectional fine-tuning, jointly optimizing wo$\to$ar and ar$\to$wo, for
10 epochs.

\paragraph{Metrics.} We report BLEU~\citep{papineni2002bleu} and
chrF++~\citep{popovic-2017-chrf} computed with sacreBLEU~\citep{post2018call},
and AfriCOMET~\citep{wang-etal-2024-afrimte}, an African-centric neural
metric built on the AfroXLM-R encoder. chrF++ is included because
character-level metrics correlate better with human judgement on
morphologically rich languages such as Wolof and Arabic.

\subsection{Results}

Table~\ref{tab:main-results} reports the full benchmark.

In the wo$\to$ar direction, all multilingual MT models benefit consistently
from fine-tuning. NLLB-200 improves by $+5.35$ BLEU and $+13.89$ chrF++. The
African-specialized AfriNLLB-12 reaches the highest lexical scores
($7.76$~BLEU, $30.72$~chrF++), with the pruned AfriNLLB-8 within $0.7$~BLEU
of the full model. The same pattern holds in ar$\to$wo: NLLB-200 improves
from $3.14$ to $7.28$~BLEU and AfriNLLB-12 reaches the best lexical
performance at $8.75$~BLEU and $33.08$~chrF++.

Two observations stand out. First, mT5-base behaves asymmetrically:
fine-tuning yields a small but measurable gain when Arabic is the target
(chrF++ rises from $0.17$ to $11.10$), but no gain when Wolof is the target.
This is consistent with mT5's pre-training mix---mC4 contains substantial
Arabic but virtually no Wolof---and suggests that the corpus is a useful
fine-tuning resource only for models that already have some Wolof coverage.
NLLB-200 and AfriNLLB satisfy this condition; mT5-base does not, despite
seeing the same training data.

Second, lexical and neural metrics do not fully agree on system ranking. AfriNLLB-12 obtains the best BLEU and chrF++ scores in both directions, while NLLB-200 fine-tuned obtains the highest AfriCOMET scores. Since the Wolof--Arabic pair is not covered by AfriCOMET's training data, we report these scores for completeness but base our comparison primarily on the lexical metrics, leaving a human evaluation of system outputs to future work.
These results demonstrate that MudawanSn provides a useful supervision signal for multilingual models that already possess some degree of Wolof representation.

%% file: sections/conclusion.tex
\section{Conclusion}
\label{sec:conclusion}

We have introduced a publicly available Wolof--Arabic parallel
corpus, a gold-standard resource of $1{,}271$ manually translated and
aligned sentence pairs drawn from Senegalese news sources. We documented the construction protocol and reported baseline machine translation experiments spanning three architectural families. Fine-tuning on our corpus produces consistent improvements over zero-shot baselines for all multilingual MT models, with African-specialized variants reaching the highest scores.
Beyond machine translation, the corpus offers a foundation for contrastive
linguistic studies of Wolof and Arabic, for the development of Wolof-aware
tokenizers and morphological analyzers, and for broader NLP research on
low-resource African language pairs involving Arabic. By releasing the corpus, splits, and code openly under a CC BY-NC license, we aim to open a line of work that has so far been entirely absent from the computational linguistics literature.

%% file: sections/limitation.tex
\section*{Limitations}
\label{sec:limitations}

\paragraph{Script coverage.}
Our corpus is limited to the Latin orthography of Wolof and therefore
does not include Wolofal. Given the scarcity of publicly available
Ajami-Wolof resources and Latin--Ajami transliteration tools, extending
MudawanSn to cover Wolofal remains an important direction for future
work. Such an extension would enable the study of the impact of shared
Arabic-derived scripts on Arabic--Wolof machine translation.

\paragraph{Corpus size.} At $1{,}271$ sentence-aligned pairs, the corpus is
modest in absolute size. This reflects the manual effort and multi-stage
quality control required to construct a gold-standard bilingual resource
from scratch for a language pair with no prior parallel data. Scaling the
corpus up and expanding it to additional domains is our primary direction
for future work.

\paragraph{Single-translator design.} The entire Arabic side was produced by a single bilingual translator. Although the workflow included dictionary-based verification, consultation with native speakers for culturally loaded items, and a semantic verification stage with human review of flagged pairs, no independent second translation or inter-translator agreement measure was collected. Translation choices therefore reflect one translator's stylistic and lexical preferences, particularly in register selection within Modern Standard Arabic.

\paragraph{Coverage of idiomatic expressions.} During translation, highly
localized or culture-specific Wolof idiomatic expressions were deliberately
avoided in favour of constructions that admit a precise mapping to Modern
Standard Arabic. This choice was made to ensure unambiguous parallel
quality and alignment robustness for machine translation training, but it
inevitably underrepresents the figurative register of Senegalese news
Wolof.

\paragraph{Domain coverage.} All source texts come from Senegalese online
news outlets, with politics and civic discourse over-represented relative
to other registers. Conversational, literary, religious, and technical
domains are entirely absent. Downstream systems fine-tuned solely on this
corpus will inherit this bias and should be expected to transfer imperfectly to other registers.